%% file: main.tex
\documentclass[letterpaper, 10 pt, conference]{ieeeconf}  %
\IEEEoverridecommandlockouts    %
\usepackage{multirow}
\usepackage{lipsum}
\usepackage{amsmath}
\usepackage{amssymb}
\usepackage[ruled,vlined, linesnumbered]{algorithm2e}
\usepackage{graphicx}
\graphicspath{{./images/}}
\usepackage{booktabs}
\usepackage{subcaption}
\usepackage{siunitx}
\usepackage{makecell}
\usepackage{multirow}
\usepackage{array}
\usepackage{graphicx}
\usepackage{calc}

\usepackage{graphicx} %
\usepackage{tikz}
 
\usepackage{pgfplots}
\pgfplotsset{compat=1.16}
\usepackage{cite}
\newcommand{\Needspace}[1]{\par\ifdim\dimexpr\pagegoal-\pagetotal\relax<#1\relax\newpage\fi}
\usepackage{microtype}      %

\usepackage{xcolor}
\usepackage{hyperref}
\hypersetup{hidelinks}

\newtheorem{proposition}{Proposition}

\title{\LARGE \bf Safe Approximate Nonlinear Model Predictive Control\\ Without Expert Demonstrations}

\author{
	\parbox{\textwidth}{%
		\centering
        Mihaela-Larisa~Clement$^{1,2}$, Mónika~Farsang$^{1}$, Johannes~Edelmann$^{3}$, Manfred~Plöchl$^{3}$,\\
        Radu~Grosu$^{1}$, Ezio~Bartocci$^{1}$, Agnes~Poks$^{3}$ and Luca~Marzari$^{1}$
	}%
	\thanks{$^{1}$Institute of Computer Engineering, TU Wien, Vienna, Austria}%
    \thanks{$^{2}$AIT Austrian Institute of Technology, Vienna, Austria}%
	\thanks{$^{3}$Institute of Mechanics and Mechatronics, TU Wien, Vienna, Austria}%
    \thanks{Corr. author: \tt\small mihaela-larisa.clement@tuwien.ac.at}%
}

\newtheorem{assumption}{Assumption}
\newtheorem{theorem}{Theorem}

\newtheorem{lemma}{Lemma}

\begin{document}
\setlength{\abovedisplayskip}{3pt plus 1pt minus 1pt}\setlength{\belowdisplayskip}{3pt plus 1pt minus 1pt}\setlength{\abovedisplayshortskip}{1pt}\setlength{\belowdisplayshortskip}{2pt}
	
\maketitle
\thispagestyle{empty}
\pagestyle{empty}

\begin{abstract}
Repeated online optimization makes nonlinear model predictive control (NMPC) computationally demanding for real-time applications. Neural approximations reduce this cost but often require large datasets of expert NMPC solutions. We introduce \textit{Sequential Approximate MPC (Seq-AMPC)}, a label-free recurrent framework that generates complete control sequences along predicted trajectories using parameters shared across the prediction horizon. The policy is trained directly through known dynamics with losses for control cost, constraint satisfaction, and a cost-decrease condition aligned with a feasible-backup safety filter. Building on this filter, we introduce a stage-cost-dependent acceptance tolerance that preserves recursive feasibility, constraint satisfaction, and asymptotic convergence under nominal MPC assumptions, independently of neural approximation error. We further derive Hoeffding-based finite-sample bounds for the expected fraction of applied proposals and closed-loop cost discrepancy relative to NMPC. We evaluate Seq-AMPC on a quadcopter, chain of masses, obstacle-avoidance vehicle, and robotic manipulator, comparing recurrent and feedforward policies under label-free and imitation training. Across four benchmarks, training takes 0.2--1.7 GPU hours per policy without expert control labels. The filtered controller applies 97.2--99.1\% of neural proposals and achieves mean closed-loop cost ratios within 0.3\% of NMPC on jointly completed episodes. Tests with suboptimal feasible initial plans further demonstrate the benefit of neural replanning.
\end{abstract}

\input{sections/introduction}
\input{sections/problem}
\input{sections/method}
\input{sections/safety}
\input{sections/evaluation}
\input{sections/conclusion}

\bibliographystyle{IEEEtran}
\bibliography{references}

\end{document}

%% file: sections/introduction.tex
\section{Introduction}\label{sec:intro}

Nonlinear model predictive control (NMPC) is widely used in safety-critical applications, including autonomous vehicles and aerial robots, as it optimizes control performance while explicitly accounting for nonlinear dynamics and operational constraints~\cite{rawlings_model_2017}. However, its deployment is often limited by the computational cost of solving a nonlinear optimization problem at every control step, particularly for high-dimensional systems, long prediction horizons, and nonconvex constraints such as obstacle avoidance. Moreover, the variable execution time of numerical solvers complicates real-time deployment on resource-constrained platforms.

\begin{figure*}[!t]
    \centering
    \includegraphics[width=\linewidth]{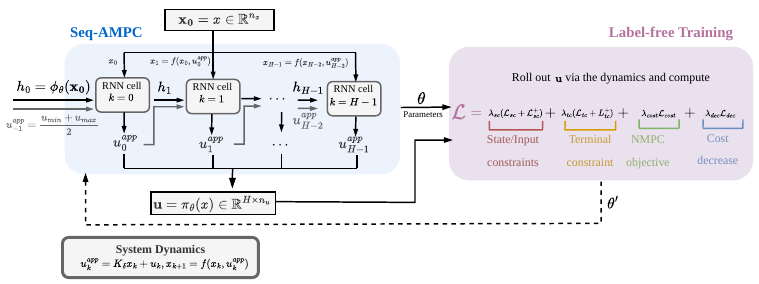}
    \caption{Overview of Seq-AMPC. A shared recurrent cell sequentially generates applied inputs and the corresponding nominal control sequence. Training evaluates model-predicted trajectories, while the runtime filter checks each proposal against a feasible backup.}
    \label{fig:Proposed_architecture}
\end{figure*}

\paragraph*{Related Work} Approximate MPC (AMPC) addresses these limitations by replacing online optimization with a neural policy trained offline to approximate the MPC control law~\cite{DBLP:conf/amcc/ChenSALKPM18,DBLP:journals/tcyb/KargL20,alsmeier2024imitation}. Existing approaches preserve theoretical guarantees by either explicitly accounting for approximation errors in the controller design and statistically certifying their magnitude~\cite{DBLP:journals/csysl/HertneckKTA18,DBLP:journals/ral/NubertKBAT20}, or deriving closed-loop guarantees under uniform approximation error bounds~\cite{DBLP:journals/csysl/PaulsonM20,tokmak2025automatic}. Nonetheless, these methods typically rely on supervised imitation learning, requiring large datasets of expert-generated solutions obtained by repeatedly solving the underlying NMPC problem. This introduces a substantial offline computational burden that becomes increasingly prohibitive for complex nonlinear systems. Guarantees based on statistical approximation bounds apply to the certified distribution with a prescribed confidence level. Neural approximation has also been applied to predictive safety filters~\cite{didier2023approximate}.

An alternative approach was introduced by Hose et al.~\cite{DBLP:journals/tcst/HoseKZT25}, who proposed a safety-augmented AMPC framework that decouples closed-loop safety guarantees from the accuracy of the learned controller. Instead of predicting a single control action, the neural network generates an entire input sequence, which is accepted only if it is feasible and its cost does not exceed that of a stored feasible backup. Otherwise, a previously feasible sequence, shifted forward and extended by a terminal controller, is applied. In the nominal setting considered here, this mechanism guarantees constraint satisfaction and convergence under the terminal and initialization assumptions, independently of the network's accuracy. The shifted-backup construction follows the suboptimal MPC principle~\cite{scokaert1999suboptimal,lazar2009predictive}. Nevertheless, the quality of the learned policy remains crucial for control performance, as it determines how frequently the proposed sequences are accepted rather than replaced by the backup. Despite these guarantees, the approach of Hose et al. relies on a feedforward multi-layer perceptron (MLP) trained through supervised imitation of NMPC solution sequences. 

This introduces two important limitations. First, predicting the entire control horizon through a feedforward architecture does not explicitly exploit the sequential structure and temporal dependencies underlying predictive control. Second, generating the required expert demonstrations involves repeatedly solving computationally expensive NMPC problems. In fact, Hose et al. report training MLPs on millions of expert-generated samples, followed by approximately 36--37 hours of GPU training per benchmark~\cite{DBLP:journals/tcst/HoseKZT25}. 
More importantly, minimizing imitation error does not necessarily produce control sequences that satisfy the runtime acceptance conditions, particularly in nonlinear environments with complex constraints. 

Recurrent neural network architectures~\cite{RNN,LSTM} offer a natural way to exploit the sequential structure and temporal dependencies of predictive control~\cite{liu2022recurrent,bonassi_recurrent_2022}. Moreover, physics-informed learning and differentiable predictive control enable neural controllers to be trained directly from dynamical models and control objectives~\cite{drgona2022dpc,Drgona2025SafePIML,Li2025RecurrentControl}. Such formulations can incorporate barrier functions~\cite{shawcortez2022dpc} or Lyapunov conditions~\cite{mukherjee2022lyapunov} for safety and stability guarantees, and model-based learning has also been applied to NMPC solver iterations~\cite{lueken2024learning}. Here, the objective is to train recurrent proposals without expert demonstrations while explicitly encouraging the feasibility and relative-cost conditions of the feasible-backup filter.

\paragraph*{Contributions} In this paper, we address these limitations by proposing \emph{Sequential Approximate MPC (Seq-AMPC)}, a physics-informed, label-free control framework combining acceptance-aligned learning, statistical performance certification, and safety-augmented deployment. As illustrated in Fig.~\ref{fig:Proposed_architecture}, our approach employs a recurrent neural policy that sequentially generates an entire control horizon using shared parameters. Rather than imitating expert solutions, we train the policy through known dynamics using the original NMPC objective and penalties for state constraints, terminal-set membership, and cost decrease. We further introduce a stage-cost-dependent acceptance tolerance and establish nominal convergence under the stated MPC assumptions, while preserving recursive feasibility and constraint satisfaction through the inherited backup mechanism. Finally, we derive Hoeffding-based finite-sample bounds on expected proposal application and closed-loop cost discrepancy relative to NMPC for fixed policies over independently sampled test episodes.

Our empirical evaluation covers four constrained benchmarks: a quadcopter, a chain of masses, an obstacle-avoidance vehicle, and a robotic manipulator. Across these tasks, the experiments compare label-free and imitation-based training, assess the effect of measurement noise, and evaluate closed-loop cost, proposal application, and complete online decision times. Tests with suboptimal feasible initializations further assess the benefit of neural replanning. Code and datasets will be released upon acceptance.

%% file: sections/problem.tex
\section{Problem Formulation}\label{sec:problem}

\subsection{Nonlinear MPC}\label{sec:system}

We adopt the safety-augmented MPC formulation of Hose et al.~\cite{DBLP:journals/tcst/HoseKZT25} in the nominal case. Consider the discrete-time system
\begin{equation}
 x_{t+1}=f(x_t,u^{\mathrm{app}}_t),\quad x_t\in\mathcal{X}\subseteq\mathbb{R}^{n_x},\ \ u^{\mathrm{app}}_t\in\mathcal{U}\subseteq\mathbb{R}^{n_u},\label{eq:dynamics}
\end{equation}
where $t\in\mathbb{N}$ is the time index, $f$ the known dynamics and $u^{\mathrm{app}}_t$ the input applied to the system. The state constraint set $\mathcal{X}$ may be nonconvex because of obstacles, the input constraint set $\mathcal{U}=[u_{\min},u_{\max}]$ is a box, and the origin is the desired equilibrium, $f(0,0)=0$. As in~\cite{DBLP:journals/tcst/HoseKZT25}, predictions use a pre-stabilizing gain $K_\delta$, so that the applied input $u^{\mathrm{app}}=K_\delta x+u$ is the sum of a state feedback and the nominal input $u$, which is the decision variable. Given the measured state $x_t$, a nominal input sequence $\mathbf{u}=[u_0^\top,\ldots,u_{H-1}^\top]^\top\in\mathbb{R}^{Hn_u}$ over the horizon $H$ generates, from $x_0=x_t$, the predicted trajectory
\begin{equation}
 u^{\mathrm{app}}_k=K_\delta x_k+u_k,\ \ x_{k+1}=f(x_k,u^{\mathrm{app}}_k),\ \ k=0,\ldots,H-1.\label{eq:prediction}
\end{equation}
The MPC cost of the state $x_t$ and the sequence $\mathbf{u}$ is
\begin{equation}
 J(x_t,\mathbf{u})=J_f(x_H)+\sum_{k=0}^{H-1}\ell(x_k,u^{\mathrm{app}}_k),\label{eq:cost}
\end{equation}
with the stage cost $\ell(x,u^{\mathrm{app}})=x^\top Qx+(u^{\mathrm{app}})^\top Ru^{\mathrm{app}}$ and the terminal cost $J_f(x)=x^\top Px$, where $Q,R,P\succ0$. With the terminal set $\mathcal{X}_f=\{x\mid x^\top Px\le\alpha^2\}$, the admissible sequences are
\begin{equation}
\mathcal{U}^H(x_t)=\{\mathbf{u}\mid x_k\in\mathcal{X},\ u^{\mathrm{app}}_k\in\mathcal{U},\ k<H,\ x_H\in\mathcal{X}_f\},\label{eq:feasset}
\end{equation}
and NMPC applies at every step the first input of a solution of $\min_{\mathbf{u}\in\mathcal{U}^H(x_t)}J(x_t,\mathbf{u})$.

\begin{assumption}[Terminal ingredients~{\cite{DBLP:journals/tcst/HoseKZT25}}]\label{ass:terminal}
There is a terminal gain $K_f$ such that, for all $x\in\mathcal{X}_f$, (i) $x\in\mathcal{X}$ and $K_fx\in\mathcal{U}$, (ii) $f(x,K_fx)\in\mathcal{X}_f$, and (iii) $J_f(f(x,K_fx))-J_f(x)\le-\ell(x,K_fx)$.
\end{assumption}

\subsection{Approximate MPC with a feasible backup}\label{sec:safety_filter}

A horizon policy $\pi_\theta:\mathbb{R}^{n_x}\to\mathbb{R}^{Hn_u}$ with parameters $\theta$ maps the measured state $x_t$ to a complete nominal input sequence, the \emph{proposal}, which approximates an NMPC solution. Since the policy does not guarantee constraint satisfaction, a safety filter evaluates the proposal against a feasible backup sequence, obtained by shifting the previously selected sequence. For a state $x$ and a sequence $\mathbf{u}$ with predicted terminal state $x_H$, the shift
\begin{equation}
 \sigma(x,\mathbf{u})=\big[u_1^\top,\ \ldots,\ u_{H-1}^\top,\ \big((K_f-K_\delta)x_H\big)^\top\big]^\top\label{eq:shift}
\end{equation}
drops the first input and appends the terminal law, so that the last applied input is $K_fx_H$.

\begin{lemma}[cf.~{\cite{scokaert1999suboptimal,DBLP:journals/tcst/HoseKZT25}}]\label{lem:shift}
Let Assumption~\ref{ass:terminal} hold, $\mathbf{u}\in\mathcal{U}^H(x)$ and $x^+=f(x,u^{\mathrm{app}}_0)$. Then $\sigma(x,\mathbf{u})\in\mathcal{U}^H(x^+)$ and $J(x^+,\sigma(x,\mathbf{u}))\le J(x,\mathbf{u})-\ell(x,u^{\mathrm{app}}_0)$.
\end{lemma}
\begin{proof}
The shift reproduces $x_1,\ldots,x_H$ from $x^+=x_1$ and appends $x_{H+1}=f(x_H,K_fx_H)$. Assumption~\ref{ass:terminal}(i) and (ii) give $K_fx_H\in\mathcal{U}$ and $x_{H+1}\in\mathcal{X}_f$, and (iii) gives $J(x^+,\sigma(x,\mathbf{u}))=J(x,\mathbf{u})-\ell(x,u^{\mathrm{app}}_0)+\ell(x_H,K_fx_H)+J_f(x_{H+1})-J_f(x_H)\le J(x,\mathbf{u})-\ell(x,u^{\mathrm{app}}_0)$.
\end{proof}

\begin{assumption}[Feasible initialization~{\cite{DBLP:journals/tcst/HoseKZT25}}]\label{ass:init}
At $t=0$, a backup sequence $\hat{\mathbf{u}}(0)\in\mathcal{U}^H(x_0)$ is available.
\end{assumption}
Assumption~\ref{ass:init} requires an initial feasible input sequence. Such a sequence may be obtained from a feasibility-checked neural proposal, a known admissible trajectory, or an available controller whose predicted trajectory satisfies the horizon and terminal constraints. An NMPC solver is therefore optional for initialization. In our experiments, initial states are sampled at random among those admitting a feasible NMPC solution, which provides a common feasible starting point for the controller comparisons. Once initialized, the nominal algorithm generates and checks neural proposals and updates the feasible backup without solving an online optimization problem.

The filter of~\cite{DBLP:journals/tcst/HoseKZT25} applies the proposal $\mathbf{u}$ only if its predicted trajectory satisfies all constraints and its cost does not exceed that of the backup $\hat{\mathbf{u}}(t)$,
\begin{equation}
 \mathbf{u}\in\mathcal{U}^H(x_t),\qquad J(x_t,\mathbf{u})\le J(x_t,\hat{\mathbf{u}}(t)),\label{eq:hose_test}
\end{equation}
and otherwise applies the backup. By Lemma~\ref{lem:shift} and induction from Assumption~\ref{ass:init}, this yields constraint satisfaction and convergence to the origin for any policy~\cite[Thm.~1]{DBLP:journals/tcst/HoseKZT25}, while the policy determines how often its proposals are applied. Near the equilibrium, however, the costs of proposal and backup both approach zero, and~\eqref{eq:hose_test} rejects proposals that differ from the backup by numerically small amounts.

%% file: sections/method.tex
\section{Sequential Label-Free Approximate NMPC}\label{sec:method}

\subsection{Recurrent horizon policy}\label{sec:arch}

To exploit the sequential structure of NMPC, Seq-AMPC generates each input from the corresponding predicted state using a recurrent cell $g_\theta$ shared across the horizon (Fig.~\ref{fig:Proposed_architecture}). Starting from $x_0=x$, a two-layer encoder initializes the hidden state as $h_0=\phi_\theta(x)$, after which the policy evolves according to
\begin{equation}\label{eq:rnn}
\begin{aligned}
h_{k+1}&=g_\theta\big(h_k,\,x_k,\,u^{\mathrm{app}}_{k-1}\big),\qquad u_k=u^{\mathrm{app}}_k-K_\delta x_k,\\
u^{\mathrm{app}}_k&=s\big(Wh_{k+1}+b+s^{-1}(K_\delta x_k)\big),
\end{aligned}
\end{equation}
for $k=0,\ldots,H-1$. At each prediction step, the $\tanh$ cell receives the normalized predicted state and the preceding applied input, initialized as $u^{\mathrm{app}}_{-1}=u_{\mathrm{mid}}=(u_{\min}+u_{\max})/2$. Its output determines the next applied input through $s(z)=u_{\min}+(u_{\max}-u_{\min})\odot\mathrm{sigmoid}(z)$, which maps elementwise onto the input box. The dynamics~\eqref{eq:prediction} then propagate this input to obtain the state used at the subsequent prediction step.

The offset $s^{-1}(K_\delta x_k)$ incorporates the pre-stabilizing feedback into this bounded output map, with $K_\delta x_k$ clipped to the interior of $\mathcal{U}$ before inversion. Consequently, a zero output layer reproduces the pre-stabilized motion $u_k=0$ whenever $K_\delta x_k$ lies in the interior of $\mathcal{U}$. The corresponding nominal inputs form the proposal $\mathbf{u}=\pi_\theta(x)=[u_0^\top,\ldots,u_{H-1}^\top]^\top$, which is used for both training and deployment. Since $g_\theta$, $W$ and $b$ are shared across steps, the parameter count is independent of $H$, allowing the same policy to be unrolled at different horizons.

\subsection{Acceptance-aligned label-free training}\label{sec:loss}

The safety filter accepts a proposed sequence according to its feasibility and cost relative to the stored backup. To align learning with these requirements, we propagate the policy outputs through the known dynamics and evaluate both constraint penalties and the MPC objective along the resulting trajectory. However, a low trajectory cost alone does not ensure acceptance, which also motivates a loss comparing consecutive proposals. The following terms define this training objective for a mini-batch of $B$ initial states $x^{(b)}$ with proposals $\mathbf{u}^{(b)}$.

\textbf{Constraint penalties.} To encourage feasibility along the predicted trajectory, we penalize violations of the state, input and terminal constraints. Writing the normalized stage constraints as $g_j(x_k,u^{\mathrm{app}}_k)\le0$ and the terminal-set residual as $\psi(x)=\sqrt{x^\top Px}/\alpha-1$, we define
\begin{equation}\label{eq:sc_tc}
\begin{aligned}
\mathcal{L}_{\mathrm{sc}}&=\frac{1}{B}\sum_{b=1}^{B}\Big[\max_{k,j}\,g_j\big(x^{(b)}_k,u^{\mathrm{app},(b)}_k\big)\Big]_+^2,\\
\mathcal{L}_{\mathrm{tc}}&=\frac{1}{B}\sum_{b=1}^{B}\big[\psi(x^{(b)}_H)\big]_+^2,
\end{aligned}
\end{equation}
where $[a]_+=\max\{0,a\}$ retains only positive residuals, so both penalties vanish when the corresponding constraints are satisfied.

\textbf{Cost and cost-decrease penalties.} Alongside feasibility, we encourage control performance by minimizing the normalized MPC cost
\begin{equation}\label{eq:cost_loss}
\mathcal{L}_{\mathrm{cost}}=\frac{1}{B}\sum_{b=1}^{B}\frac{J(x^{(b)},\mathbf{u}^{(b)})}{\max\{\mathrm{sg}[J(x^{(b)},\mathbf{0})],J_{\min}\}},
\end{equation}
where normalization by the cost of the pre-stabilized motion reduces differences in loss scale between states near and far from the equilibrium. The stop-gradient $\mathrm{sg}[\cdot]$ holds this normalization fixed during differentiation, while $J_{\min}>0$ prevents division by zero.

To additionally account for the filter's relative-cost condition, we compare the proposal at the successor state $x^+=f(x,u^{\mathrm{app}}_0)$ with the shifted current proposal $\hat{\mathbf{u}}^+=\sigma(x,\mathbf{u})$. This shifted sequence becomes the backup at $x^+$ if $\mathbf{u}$ is applied, providing the training reference for the next proposal. Denoting their costs by $J^+_\theta=J(x^+,\pi_\theta(x^+))$ and $\hat J^+=J(x^+,\hat{\mathbf{u}}^+)$, respectively, we define the cost-decrease loss as
\begin{equation}\label{eq:dec_loss}
\mathcal{L}_{\mathrm{dec}}=\frac{1}{B}\sum_{b=1}^{B}e^{(b)}\bigg[\frac{J^{+(b)}_\theta-\hat J^{+(b)}}{\max\{\hat J^{+(b)},J_{\min}\}}+\mu\bigg]_+,
\end{equation}
where $e^{(b)}\in\{0,1\}$ restricts the comparison to samples with admissible current and shifted proposals, and $\mu>0$ introduces a margin for improvement. The loss thus encourages the successor proposal to improve on the shifted current proposal. Nevertheless, the stored backup depends on the sequence actually selected by the filter, so this training comparison promotes acceptance while leaving its verification to the runtime filter.

\textbf{Training objective.} Combining these terms, we train the policy with the weighted objective
\begin{equation}\label{eq:total_loss}
\mathcal{L}=\lambda_{\mathrm{sc}}(\mathcal{L}_{\mathrm{sc}}+\mathcal{L}^{+}_{\mathrm{sc}})+\lambda_{\mathrm{tc}}(\mathcal{L}_{\mathrm{tc}}+\mathcal{L}^{+}_{\mathrm{tc}})+\lambda_{\mathrm{cost}}\mathcal{L}_{\mathrm{cost}}+\lambda_{\mathrm{dec}}\mathcal{L}_{\mathrm{dec}},
\end{equation}
where the coefficients $\lambda$ balance the loss terms. To encourage feasibility of the successor proposal as well as its relative cost, $\mathcal{L}^{+}_{\mathrm{sc}}$ and $\mathcal{L}^{+}_{\mathrm{tc}}$ also evaluate the constraint penalties on $\pi_\theta(x^+)$ for eligible samples.

To further improve performance near the equilibrium, we subsequently fine-tune the trained policy by minimizing the terminal-cost decrease penalty
\begin{equation}\label{eq:finetune}
\mathcal{L}_{\mathrm{local}}=\mathbb{E}\left[\frac{\big[\ell(x,u^{\mathrm{app}})+J_f(f(x,u^{\mathrm{app}}))-J_f(x)\big]_+}{J_f(x)+\nu}\right],
\end{equation}
with a small constant $\nu>0$. Retaining~\eqref{eq:total_loss} at reduced weight during this refinement preserves the broader training objectives while emphasizing local terminal-cost decrease. Both training stages therefore rely on the known dynamics and control objective without expert demonstrations.

\textbf{Training data.} We sample 130k initial states from the operating region and retain those for which NMPC finds a feasible solution. Both label-free and imitation policies use these sampled initial states, but optimize different objectives. Whereas imitation fits the corresponding NMPC solution sequences, label-free training evaluates predicted trajectories and successor-state losses through~\eqref{eq:total_loss}, without collecting additional training states. Consistent with the intended runtime behavior, the final model is selected by proposal feasibility and acceptance on held-out validation states.

%% file: sections/safety.tex
\Needspace{12\baselineskip}
\section{Safety-Augmented Control and Guarantees}\label{sec:safety}

\subsection{Recursive feasibility and convergence}\label{sec:theory}

Although the training objective encourages feasible proposals with low cost, it does not guarantee that these properties hold at every control step. Consequently, deployment requires explicit verification of each proposal $\mathbf{u}=\pi_\theta(x_t)$ against the stored feasible backup $\hat{\mathbf{u}}(t)$. To allow approximate solutions within this mechanism, we extend the strict cost condition~\eqref{eq:hose_test} and accept the proposal if
\begin{equation}
 \mathbf{u}\in\mathcal{U}^H(x_t),\qquad J(x_t,\mathbf{u})\le J(x_t,\hat{\mathbf{u}}(t))+\epsilon_t.
 \label{eq:acceptance}
\end{equation}
Otherwise, the backup is applied, so the selected sequence remains feasible in either case. To also retain convergence under the relaxed cost condition, we couple the admissible tolerance to the preceding stage cost through
\begin{equation}
 \epsilon_{t+1}=\min\{\bar\epsilon,\ \eta\,\ell(x_t,u^{\mathrm{app}}_t)\},\qquad \epsilon_0=\bar\epsilon,\label{eq:adaptive_tol}
\end{equation}
where $\bar\epsilon\ge0$ caps the relaxation and $0\le\eta<1$ limits it to a fraction of the preceding stage cost. Thus, $\epsilon_t=0$ recovers the strict test~\eqref{eq:hose_test}, while positive tolerances allow a controlled increase relative to the backup cost. After either sequence is selected, the shift~\eqref{eq:shift} provides the backup for the next step. Algorithm~\ref{alg:filter} summarizes this procedure, whose nominal guarantees are established below.

\begin{algorithm}[t]
\SetAlgoLined
\DontPrintSemicolon
\caption{Safe Seq-AMPC (one episode)}\label{alg:filter}
\KwIn{$x_0$, $\hat{\mathbf{u}}(0)\in\mathcal{U}^H(x_0)$, policy $\pi_\theta$, $\bar\epsilon$, $\eta$}
$\epsilon_0\gets\bar\epsilon$\;
\For{$t=0,1,\ldots$}{
  $\mathbf{u}\gets\pi_\theta(x_t)$ via~\eqref{eq:rnn}\;
  \lIf{\eqref{eq:acceptance} holds}{$\tilde{\mathbf{u}}(t)\gets\mathbf{u}$}
  \lElse{$\tilde{\mathbf{u}}(t)\gets\hat{\mathbf{u}}(t)$}
  apply $u^{\mathrm{app}}_t=K_\delta x_t+\tilde{u}_0(t)$ and measure $x_{t+1}$\;
  $\hat{\mathbf{u}}(t+1)\gets\sigma(x_t,\tilde{\mathbf{u}}(t))$, $\epsilon_{t+1}$ by~\eqref{eq:adaptive_tol}\;
}
\end{algorithm}

\begin{theorem}[Closed-loop guarantees]\label{thm:main}
Let Assumptions~\ref{ass:terminal} and~\ref{ass:init} hold and $x_{t+1}=f(x_t,u^{\mathrm{app}}_t)$. Then the closed loop under~\eqref{eq:acceptance} and~\eqref{eq:adaptive_tol} satisfies, for any neural policy, with $V_t=J(x_t,\tilde{\mathbf{u}}(t))$ and $\ell_t=\ell(x_t,u^{\mathrm{app}}_t)$:
(i) recursive feasibility and constraint satisfaction, $\hat{\mathbf{u}}(t),\tilde{\mathbf{u}}(t)\in\mathcal{U}^H(x_t)$, $x_t\in\mathcal{X}$ and $u^{\mathrm{app}}_t\in\mathcal{U}$ for all $t\ge0$,
(ii) cost decrease, $V_{t+1}\le V_t-(1-\eta)\ell_t$ for all $t\ge0$, and
(iii) asymptotic convergence, $x_t\to0$ as $t\to\infty$.
\end{theorem}
\begin{proof}
(i) follows by induction from Assumption~\ref{ass:init} and Lemma~\ref{lem:shift}, since the filter selects either the admissible backup or an admissible proposal, and the constraints at $k=0$ in~\eqref{eq:feasset} are those on $(x_t,u^{\mathrm{app}}_t)$. (ii) By Lemma~\ref{lem:shift} and the acceptance condition,
\begin{equation*}
V_{t+1}\le J(x_{t+1},\hat{\mathbf{u}}(t+1))+\epsilon_{t+1}\le V_t-\ell_t+\eta\ell_t.
\end{equation*}
(iii) Since $V_t\ge0$, summing (ii) yields $\sum_{t=0}^{\infty}\ell_t\le V_0/(1-\eta)<\infty$, so $\ell_t\to0$, and $\ell_t\ge\lambda_{\min}(Q)\|x_t\|^2$ implies $x_t\to0$.
\end{proof}

Theorem~\ref{thm:main} separates the roles of the backup and the adaptive tolerance. Recursive feasibility and constraint satisfaction follow from the feasible-backup construction, as in~\cite[Thm.~1]{DBLP:journals/tcst/HoseKZT25}. The stage-cost dependence of the tolerance additionally preserves sufficient cost decrease for asymptotic convergence under Assumptions~\ref{ass:terminal} and~\ref{ass:init}.

Application of this result therefore depends on the terminal ingredients. Assumption~\ref{ass:terminal} holds by construction for the linear vehicle model and, for the robot arm, through a certified bound on the terminal-set radius of each sampled target that ensures satisfaction of the collision constraints. In contrast, the nonlinear quadcopter and chain of masses use numerical assessment on dense terminal-set samples, so their guarantees remain conditional on Assumption~\ref{ass:terminal}.

\subsection{Statistical performance certification}\label{sec:cert}

Theorem~\ref{thm:main} guarantees nominal feasibility and convergence independently of the learned policy, but does not quantify how frequently its proposals are applied or how its closed-loop cost compares with NMPC. We therefore complement the theorem with statistical bounds on these two quantities. Let $\mathcal{D}$ denote the distribution of feasible initial states introduced in Section~\ref{sec:safety_filter}. With the policy and reference NMPC solver settings fixed, we draw $n$ initial states independently from $\mathcal{D}$, independently of training and model selection, and evaluate both controllers over episodes of length $T$. For episode $i$, define
\begin{equation}\label{eq:metrics}
\rho_i=\frac1T\sum_{t=0}^{T-1}a_i(t),\qquad \Delta_i=\min\Big\{1,\frac{|C_i-C^{\mathrm{ref}}_i|}{\max\{C_i,C^{\mathrm{ref}}_i\}}\Big\},
\end{equation}
where $a_i(t)\in\{0,1\}$ indicates whether the neural proposal is applied at time $t$, and $C_i$ and $C^{\mathrm{ref}}_i$ are the accumulated closed-loop stage costs of Safe Seq-AMPC and NMPC, respectively. Since the reference NMPC need not attain the lowest accumulated cost, the absolute difference measures discrepancies in either direction. To include unsuccessful episodes in the assessment, we set $\Delta_i=1$ when either closed loop fails to complete and $a_i(t)=0$ after an early termination. For completed episodes with both costs zero, we set $\Delta_i=0$. These conventions keep both metrics in $[0,1]$, allowing the following finite-sample bounds.

\begin{proposition}[Statistical certification]\label{prop:cert}
Let $\bar\rho$ and $\bar\Delta$ be the empirical means of~\eqref{eq:metrics} over $n$ independent episodes. For any $\delta\in(0,1)$, with probability at least $1-\delta$,
\begin{equation}\label{eq:hoeffding}
\mathbb{E}_{\mathcal{D}}[\rho]\ge\bar\rho-r_n(\delta),\qquad\mathbb{E}_{\mathcal{D}}[\Delta]\le\bar\Delta+r_n(\delta),
\end{equation}
where $r_n(\delta)=\sqrt{\ln(2/\delta)/(2n)}$.
\end{proposition}
\begin{proof}
The policy and the solver settings are fixed before the test episodes are drawn, so $\rho_i$ and $\Delta_i$ are independent and identically distributed in $[0,1]$. Applying the one-sided Hoeffding inequality~\cite{hoeffding1963probability} to each empirical mean with failure probability $\delta/2$, followed by the union bound, establishes the result.
\end{proof}

Consequently, a prescribed accuracy $r_n(\delta)\le\gamma$ is attained with $n\ge\lceil\ln(2/\delta)/(2\gamma^2)\rceil$ episodes. This sample requirement connects the desired statistical precision to the evaluation budget, while the bounds characterize expected proposal application and cost discrepancy under $\mathcal{D}$. They therefore complement the nominal guarantees of Theorem~\ref{thm:main} with a distribution-dependent assessment of the fixed policy's performance.

%% file: sections/evaluation.tex
\section{Numerical Evaluation}\label{sec:evaluation}

\begin{figure*}[!t]
    \centering
    \includegraphics[width=0.93\linewidth]{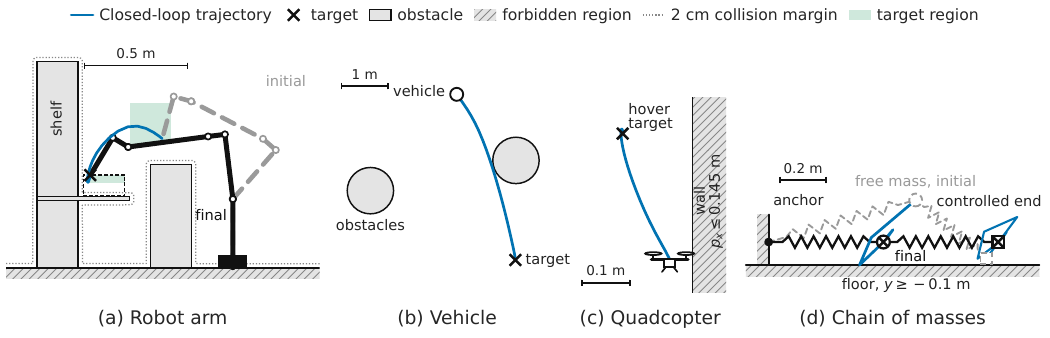}
    \caption{Benchmarks with representative Safe Seq-AMPC trajectories: (a) robot arm reaching inside a shelf, (b) vehicle among obstacles, (c) quadcopter near a wall, (d) chain of masses above a floor.}
    \label{fig:tasks}
\end{figure*}

We evaluate whether label-free recurrent proposals can improve closed-loop cost while satisfying the safety filter's acceptance condition. Four constrained benchmarks test this question against NMPC, backup-only control, and neural policies with different architectures and training objectives.

\subsection{Experimental setup}\label{sec:setup}

We consider a quadcopter, a chain of masses regulated above a floor, a planar vehicle navigating around obstacles, and a Franka manipulator performing collision-constrained reaching inside a shelf (Fig.~\ref{fig:tasks}), with state and input dimensions and prediction horizons $(n_x,n_u,H)$ of $(10,3,10)$, $(9,3,10)$, $(4,2,20)$ and $(14,7,20)$, respectively. The quadcopter and robot-arm benchmarks follow~\cite{DBLP:journals/tcst/HoseKZT25}. The vehicle is a planar double integrator subject to state, input and obstacle constraints. The robot-arm controller operates in error coordinates relative to an inverse-kinematics goal configuration, and targets are restricted to collision-free goal configurations that admit the required terminal set.

The proposed recurrent policy and a six-layer MLP, both of width 256, are trained either with the label-free objective~\eqref{eq:total_loss} or by imitation of NMPC solution sequences using mean absolute error. Both training strategies use the same 130k sampled initial states. The recurrent policy generates controls sequentially along model-predicted states, while label-free training evaluates predicted trajectories and successor-state losses. For the arm, target configurations are supplied as additional policy inputs, and the sampled initial states have zero joint velocities. Training hyperparameters were selected independently for each architecture and training objective with CARBS~\cite{fetterman2023carbs} on validation data. Training required 0.2 to 1.7 GPU hours per policy on one NVIDIA RTX PRO 6000. NMPC is implemented in acados with at most 200 SQP iterations.

Three independently trained Seq-AMPC policies per benchmark are evaluated from test initial states sampled independently of the training and validation sets. All controllers start from identical initial states, the nominal prediction model is used as the plant, and each episode comprises at most $T=100$ decisions. An episode succeeds if its final state satisfies $\|x\|\le0.05$ for the quadcopter and chain, $\|x\|\le0.1$ for the vehicle, or an end-effector error below 5\,cm and joint speeds below 0.05\,rad/s for the arm. Episodes end early on a constraint violation or solver failure. Safe Seq-AMPC and backup-only control start from one feasible NMPC solution per episode. If both the proposal and backup are inadmissible, a case excluded under Theorem~\ref{thm:main}, the implementation attempts one NMPC solve and otherwise records failure. We use $\bar\epsilon=0.01$ and $\eta=0.9$ to allow approximate proposals while reducing the admissible cost difference according to the preceding stage cost. Decision times are measured for individual control decisions on one AMD EPYC 9654 CPU core.

\input{sections/tab_success_rates}

\subsection{Closed-loop performance and statistical certification}\label{sec:main}

If the learned proposals are useful, the filter should apply most of them while retaining NMPC cost performance. Table~\ref{tab:main} supports this expectation: proposal application is high and Safe Seq-AMPC achieves near-NMPC cost with full success on all benchmarks. Excess-cost ratios are computed only over episodes completed by both compared controllers. Every Safe Seq-AMPC episode satisfies the constraints and reaches its target set without a fallback NMPC solve.

With an NMPC-initialized backup, backup-only control is already strong, with excess cost ranging from 0 to 1.59\,\% relative to NMPC. This leaves limited room for improvement and makes initial plan quality central to the comparison. To test replanning from a suboptimal feasible initialization, we compute the shared initial plan using an input penalty scaled by 100. The filter and all reported costs use the original MPC objective. Table~\ref{tab:backup_init} shows that Safe Seq-AMPC substantially reduces the excess cost of this initial plan, with more than 99\,\% of proposals applied. Thus neural replanning can improve a feasible but suboptimal plan without repeated online NMPC optimization.

\input{sections/tab_backup_init}

To quantify the statistical confidence in the observed proposal utilization and cost performance, we apply Proposition~\ref{prop:cert} to 3000 independent test episodes per fixed policy. This assessment concerns expected performance over the test-state distribution, complementing the nominal feasibility and convergence guarantees of Theorem~\ref{thm:main}. Setting $\delta=0.05$ gives a Hoeffding radius of $r_n=0.0248$, meeting the prescribed accuracy $\gamma=0.025$.

\begin{figure}[!t]
    \centering
    \includegraphics[width=\linewidth]{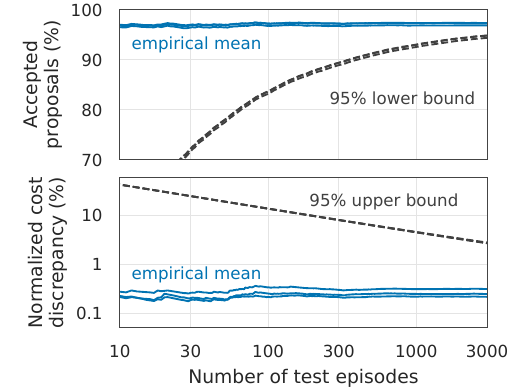}
    \caption{Empirical estimates and 95\,\% Hoeffding bounds for fixed robot-arm policies, pointwise in sample size.}
    \label{fig:cert}
\end{figure}

Fig.~\ref{fig:cert} illustrates this assessment on the robot arm. For each fixed policy, the bounds certify an expected application rate of at least 94.3\,\% and an expected normalized cost discrepancy of at most 2.8\,\%. Thus, even after accounting for finite-sample uncertainty, the learned policies retain high expected proposal utilization and a small expected cost discrepancy relative to NMPC. The two bounds hold jointly with 95\,\% confidence for each fixed policy and sample size, without asserting simultaneous coverage across policies or along the plotted curves.

\subsection{Recurrent architecture and label-free learning}\label{sec:ablation}

We compare architectures and training objectives to assess whether recurrent sequence generation and model-based label-free optimization yield effective unfiltered controllers. For each benchmark, CARBS tunes each architecture--objective combination separately, including the learning rate and number of training epochs. Table~\ref{tab:ablation} reports test performance for the best configurations found on validation data.

Under these individually tuned configurations, success is similar across methods on the quadcopter and chain, with larger differences on the vehicle and arm. On the arm, label-free Seq-AMPC has higher mean success and lower variability across policies than the label-free MLP, while both imitation architectures fail to reach the target. Both methods use the same sampled initial states, but their learning signals differ: imitation fits NMPC solution sequences, whereas label-free training propagates losses through predicted trajectories and successor-state evaluations. Since arm training states have zero joint velocities, imitation receives no expert labels for starting a horizon from a moving state. The arm results favor the recurrent label-free formulation, while the similar performance on the quadcopter and chain limits claims of a general advantage. The comparison therefore concerns the best configurations found for each method and reflects both their learning signals and their individually tuned optimization settings.

\input{sections/tab_ablation_v8}

We further assess the sensitivity of the training objective to its loss weights on the robot arm.\footnote{Across the CARBS trials, the loss weights varied over several orders of magnitude, with validation feasibility between 95\,\% and 100\,\%.} Scaling individual constraint and cost weights by factors of 0.1 and 10 preserves proposal feasibility above 99.7\,\% and changes the mean stage cost by less than 1\,\%, indicating limited sensitivity within the tested ranges.

\subsection{Safety filter and adaptive tolerance}\label{sec:raw}

The safety filter addresses constraint violations in unfiltered learned policies, which occur in 0.6 to 7.0\,\% of episodes. No violations occur with Safe Seq-AMPC in the nominal evaluation.

\begin{figure}[!t]
    \centering
    \includegraphics[width=\linewidth]{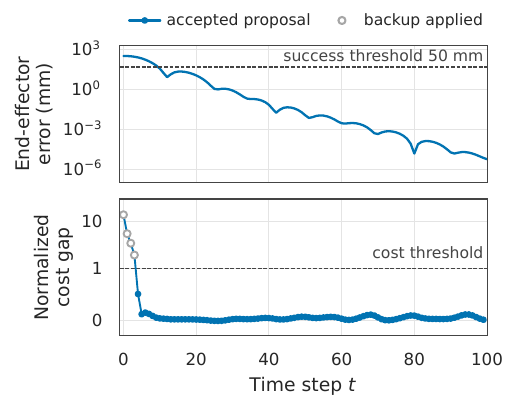}
    \caption{Robot-arm filter decisions. For $\epsilon_t>0$, the lower panel shows $[J(x_t,\mathbf{u})-J(x_t,\hat{\mathbf{u}}(t))]/\epsilon_t$. A feasible proposal passes the cost test at values at most one. When $\epsilon_t=0$, the ratio is undefined and the unnormalized cost test applies.}
    \label{fig:episode}
\end{figure}

The strict cost test~\eqref{eq:hose_test} rejects many proposals, particularly on the vehicle and arm. Over 300 test episodes, the adaptive tolerance raises application from 35.7 to 98.3\,\% on the vehicle and from 4.4 to 97.1\,\% on the arm. Crucially, the stage-cost-dependent tolerance preserves the sufficient cost-decrease condition and nominal convergence result of Theorem~\ref{thm:main}. A constant positive tolerance can yield similar empirical application rates, but does not provide the same convergence condition. Fig.~\ref{fig:episode} illustrates the normalized cost gap and threshold for one arm episode.

\subsection{Computational performance}\label{sec:horizon}

\begin{figure}[!t]
    \centering
    \includegraphics[width=\linewidth]{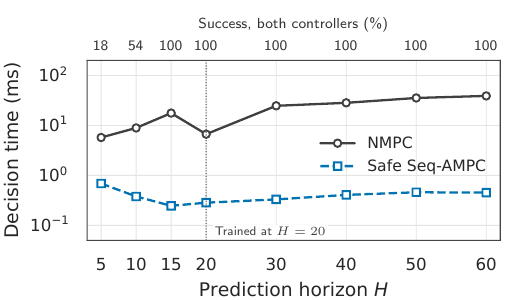}
    \caption{Decision time versus prediction horizon for an arm policy trained at $H=20$. This separate experiment uses single-precision inference and reports medians over 300 episodes on one CPU core. Safe Seq-AMPC includes inference and filter evaluation, while NMPC time is the solver time.}
    \label{fig:horizon}
\end{figure}

Table~\ref{tab:main} compares complete online decisions. Backup-only takes 0.08 to 0.13\,ms per decision because it shifts the selected plan without generating and evaluating a new candidate. Safe Seq-AMPC takes 0.32 to 0.46\,ms per decision, including 0.16 to 0.24\,ms for inference, while NMPC solve times average 12 to 35\,ms. Safe Seq-AMPC therefore adds modest computation while enabling replanning from a suboptimal feasible plan without repeated NMPC solves. 

The horizon-scaling experiment uses a separate timing protocol: for the arm policy trained at $H=20$, Safe Seq-AMPC takes 0.25 to 0.69\,ms at horizons from 5 to 60, while NMPC time grows from 5.7 to 39\,ms. Both controllers succeed less often at short horizons (Fig.~\ref{fig:horizon}).

\subsection{Measurement noise}\label{sec:noise}

\begin{figure}[!t]
    \centering
    \includegraphics[width=\linewidth]{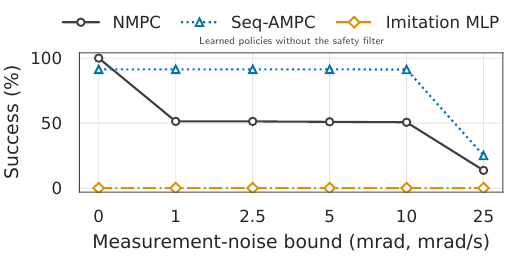}
    \caption{Robot-arm success under bounded uniform noise on joint positions (mrad) and velocities (mrad/s). Each level uses 300 episodes per policy with the success criterion of Section~\ref{sec:setup}. Learned-policy results are averaged over three policies.}
    \label{fig:noise}
\end{figure}

Fig.~\ref{fig:noise} reports unfiltered-policy success under bounded measurement noise on the robot arm, which has the largest state dimension and collision constraints. Seq-AMPC retains its nominal 91.3\,\% success up to noise bounds of 10\,mrad in joint positions and 10\,mrad/s in joint velocities. NMPC loses 49 percentage points at bounds of 1\,mrad and 1\,mrad/s. These empirical results do not extend the nominal guarantee to uncertain measurements.

%% file: sections/tab_success_rates.tex
\begin{table*}[!t]
    \centering
    \caption{Closed-loop results over 3000 test episodes per policy. Entries are mean $\pm$ standard deviation across episodes, except decision times, which are summarized across individual decisions. Excess cost is the per-episode ratio to NMPC minus one, computed on episodes completed by both controllers. Online times exclude the initial NMPC solve for Safe Seq-AMPC and Backup-only control. NMPC solve statistics use 300 episodes.}
    \label{tab:main}
    \footnotesize
    \setlength{\tabcolsep}{3pt}
    \begin{tabular}{l c c c c c c c c c}
        \toprule
        & \multicolumn{3}{c}{Success (\%)} & & \multicolumn{2}{c}{Excess cost (\%)} & \multicolumn{3}{c}{Time per decision (ms)} \\
        \cmidrule(lr){2-4} \cmidrule(lr){6-7} \cmidrule(lr){8-10}
        Benchmark & \makecell{Safe\\Seq-AMPC} & \makecell{Backup-only} & NMPC & \makecell{Proposal\\applied (\%)} & \makecell{Safe\\Seq-AMPC} & \makecell{Backup-only} & \makecell{Safe\\Seq-AMPC} & \makecell{Backup-only} & NMPC \\
        \midrule
        Quadcopter & 100.0 & 100.0 & 100.0 & 99.1\,{\scriptsize$\pm$\,0.7} & 0.10\,{\scriptsize$\pm$\,0.21} & 1.59\,{\scriptsize$\pm$\,1.17} & 0.360\,{\scriptsize$\pm$\,0.014} & 0.092\,{\scriptsize$\pm$\,0.004} & 21.6\,{\scriptsize$\pm$\,34.8} \\
        Chain of masses & 100.0 & 100.0 & 99.5 & 98.7\,{\scriptsize$\pm$\,1.0} & 0.02\,{\scriptsize$\pm$\,0.02} & 0.05\,{\scriptsize$\pm$\,0.06} & 0.321\,{\scriptsize$\pm$\,0.009} & 0.082\,{\scriptsize$\pm$\,0.003} & 29.0\,{\scriptsize$\pm$\,70.4} \\
        Vehicle & 100.0 & 100.0 & 100.0 & 98.3\,{\scriptsize$\pm$\,2.2} & 0.02\,{\scriptsize$\pm$\,0.01} & 0.04\,{\scriptsize$\pm$\,0.02} & 0.456\,{\scriptsize$\pm$\,0.012} & 0.108\,{\scriptsize$\pm$\,0.003} & 12.2\,{\scriptsize$\pm$\,29.0} \\
        Robot arm & 100.0 & 100.0 & 100.0 & 97.2\,{\scriptsize$\pm$\,2.5} & 0.26\,{\scriptsize$\pm$\,0.23} & 0.00\,{\scriptsize$\pm$\,0.00} & 0.383\,{\scriptsize$\pm$\,0.012} & 0.125\,{\scriptsize$\pm$\,0.012} & 35.3\,{\scriptsize$\pm$\,270.2} \\
        \bottomrule
    \end{tabular}
\end{table*}

%% file: sections/tab_backup_init.tex
\begin{table}[!htbp]
    \centering
    \caption{Excess closed-loop cost (\%) relative to NMPC over 300 test episodes. Both controllers start from the same feasible plan computed with a larger input penalty, and costs use the original MPC objective. Entries are mean $\pm$ standard deviation over episodes,  over three Safe Seq-AMPC policies.}
    \label{tab:backup_init}
    \footnotesize
    \setlength{\tabcolsep}{5pt}
    \begin{tabular}{l c c}
        \toprule
        Benchmark & Backup-only & Safe Seq-AMPC \\
        \midrule
        Quadcopter      & 11.2\,{\scriptsize$\pm$\,12.4} & \textbf{0.85}\,{\scriptsize$\pm$\,2.25} \\
        Chain of masses & 70.6\,{\scriptsize$\pm$\,25.5} & \textbf{0.88}\,{\scriptsize$\pm$\,2.54} \\
        Vehicle         & 18.0\,{\scriptsize$\pm$\,11.2} & \textbf{0.10}\,{\scriptsize$\pm$\,0.41} \\
        Robot arm       & 71.3\,{\scriptsize$\pm$\,22.1} & \textbf{5.3}\,{\scriptsize$\pm$\,9.7} \\
        \bottomrule
    \end{tabular}
\end{table}

%% file: sections/tab_ablation_v8.tex
\begin{table}[!htbp]
    \centering
    \caption{Success (\%) without the safety filter for the best validation-selected CARBS configurations, using 300 test episodes per policy. Entries are mean $\pm$ standard deviation across trained policies. LF: label-free training with~\eqref{eq:total_loss}. IL: imitation of NMPC solutions.}
    \label{tab:ablation}
    \footnotesize
    \setlength{\tabcolsep}{3pt}
    \begin{tabular}{l c c c c}
        \toprule
        Benchmark & \makecell{Seq-AMPC\\(LF)} & \makecell{MLP\\(LF)} & \makecell{Seq-AMPC\\(IL)} & \makecell{MLP\\(IL)} \\
        \midrule
        Quadcopter & \textbf{99.3}\,{\scriptsize$\pm$\,0.0} & 98.6\,{\scriptsize$\pm$\,0.2} & 99.3\,{\scriptsize$\pm$\,0.0} & 98.9\,{\scriptsize$\pm$\,0.4} \\
        Chain of masses & \textbf{98.7}\,{\scriptsize$\pm$\,0.0} & 98.7\,{\scriptsize$\pm$\,0.0} & 98.3\,{\scriptsize$\pm$\,0.0} & 98.3\,{\scriptsize$\pm$\,0.0} \\
        Vehicle & \textbf{96.9}\,{\scriptsize$\pm$\,0.2} & 96.2\,{\scriptsize$\pm$\,0.2} & 82.8\,{\scriptsize$\pm$\,4.5} & 94.1\,{\scriptsize$\pm$\,0.2} \\
        Robot arm & \textbf{91.3}\,{\scriptsize$\pm$\,4.0} & 76.0\,{\scriptsize$\pm$\,22.5} & 0.0\,{\scriptsize$\pm$\,0.0} & 0.0\,{\scriptsize$\pm$\,0.0} \\
        \bottomrule
    \end{tabular}
\end{table}

%% file: sections/conclusion.tex
\Needspace{18\baselineskip}
\section{Conclusion}\label{sec:conclusion}

We introduced Seq-AMPC, which generates complete control sequences recurrently and trains through known dynamics without expert labels. Its feasible-backup filter uses a stage-cost-dependent acceptance tolerance. Under the stated nominal MPC assumptions, we proved recursive feasibility, constraint satisfaction, and asymptotic convergence, and derived finite-sample bounds on expected proposal application and closed-loop cost discrepancy relative to NMPC. Experiments on four constrained benchmarks showed near-NMPC performance with reduced online computation. Future work will examine deployment on physical platforms such as mobile robots. The experiments establish that recurrent neural replanning can improve suboptimal feasible initial sequences at considerably lower online cost than NMPC.